# Mixed batches and symmetric discriminators for GAN training

Thomas Lucas*[1]   Corentin Tallec*[2]   Jakob Verbeek[1]   Yann Ollivier[3]

## Abstract

Generative adversarial networks (GANs) are powerful generative models based on providing feedback to a generative network via a discriminator network. However, the discriminator usually assesses individual samples. This prevents the discriminator from accessing global distributional statistics of generated samples, and often leads to *mode dropping*: the generator models only part of the target distribution. We propose to feed the discriminator with *mixed batches* of true and fake samples, and train it to predict the ratio of true samples in the batch. The latter score does not depend on the order of samples in a batch. Rather than learning this invariance, we introduce a generic permutation-invariant discriminator architecture. This architecture is provably a universal approximator of all symmetric functions. Experimentally, our approach reduces mode collapse in GANs on two synthetic datasets, and obtains good results on the CIFAR10 and CelebA datasets, both qualitatively and quantitatively.

## 1. Introduction

Estimating generative models from unlabeled data is one of the challenges in unsupervised learning. Recently, several latent variable approaches have been proposed to learn flexible density estimators together with efficient sampling, such as generative adversarial networks (GANs) (Goodfellow et al., 2014), variational autoencoders (Kingma & Welling, 2014; Rezende et al., 2014), iterative transformation of noise (Sohl-Dickstein et al., 2015), or non-volume preserving transformations (Dinh et al., 2017).

In this work we focus on GANs, currently the most convincing source of samples of natural images (Karras et al., 2018). GANs consist of a generator and a discriminator network. The generator maps samples from a latent random variable with a basic prior, such as a multi-variate Gaussian, to the observation space. This defines a probability distribution over the observation space. A discriminator network is trained to distinguish between generated samples and true samples in the observation space. The generator, on the other hand, is trained to fool the discriminator. In an idealized setting with unbounded capacity of both networks and infinite training data, the generator should converge to the distribution from which the training data has been sampled.

In most adversarial setups, the discriminator classifies individual data samples. Consequently, it cannot directly detect discrepancies between the *distribution* of generated samples and global statistics of the training distribution, such as its moments or quantiles. For instance, if the generator models a restricted part of the support of the target distribution very well, this can fool the discriminator at the level of individual samples, a phenomenon known as *mode dropping*. In such a case there is little incentive for the generator to model other parts of the support of the target distribution. A more thorough explanation of this effect can be found in (Salimans et al., 2016).

In order to access global distributional statistics, imagine a discriminator that could somehow take full probability distributions as its input. This is impossible in practice. Still, it is possible to feed large batches of training or generated samples to the discriminator, as an approximation of the corresponding distributions. The discriminator can compute statistics on those batches and detect discrepancies between the two distributions. For instance, if a large batch exhibits only one mode from a multimodal distribution, the discriminator would notice the discrepancy right away. Even though a single batch may not encompass all modes of the distribution, it will still convey more information about missing modes than an individual example.

Training the discriminator to discriminate "pure" batches with only real or only synthetic samples makes its task too easy, as a single bad sample reveals the whole batch as synthetic. Instead, we introduce a "mixed" batch discrimination task in which the discriminator needs to predict the ratio of real samples in a batch.

*Equal contribution  [1]Université Grenoble Alpes, Inria, CNRS, Grenoble INP, LJK, 38000 Grenoble, France.  [2]Université Paris Sud, INRIA, équipe TAU, Gif-sur-Yvette, 91190, France.  [3]Facebook Artificial Intelligence Research Paris, France. Correspondence to: Corentin Tallec <corentin.tallec@inria.fr>, Thomas Lucas <thomas.lucas@inria.fr>.

This use of batches differs from traditional minibatch learning. The batch is not used as a computational trick to increase parallelism, but as an approximate distribution, on which to compute global statistics.

A naive way of doing so would be to concatenate the samples in the batch, feeding the discriminator a single tensor containing all the samples. However, this is parameter-hungry, and the computed statistics are not automatically invariant to the order of samples in the batch. To compute functions that depend on the samples only through their distribution, it is necessary to restrict the class of discriminator networks to *permutation-invariant* functions of the batch. For this, we adapt and extend an architecture from McGregor (2007) to compute symmetric functions of the input. We show this can be done with minimal modification to existing architectures, at a negligible computational overhead w.r.t. ordinary batch processing.

In summary, our contributions are the following:

- Naively training the discriminator to discriminate "pure" batches with only real or only synthetic samples makes its task way too easy. We introduce a discrimination loss based on *mixed* batches of true and fake samples, that avoids this pitfall. We derive the associated optimal discriminator.
- We provide a principled way of defining neural networks that are permutation-invariant over a batch of samples. We formally prove that the resulting class of functions comprises all symmetric continuous functions, and only symmetric functions.
- We apply these insights to GANs, with good experimental results, both qualitatively and quantitatively.

We believe that discriminating between distributions at the batch level provides an equally principled alternative to approaches to GANs based on duality formulas (Nowozin et al., 2016; Gulrajani et al., 2017; Arjovsky et al., 2017).

## 2. Related work

The training of generative models via distributional rather than pointwise information has been explored in several recent contributions. Batch discrimination (Salimans et al., 2016) uses a handmade layer to compute batch statistics which are then combined with sample-specific features to enhance individual sample discrimination. Karras et al. (2018) directly compute the standard deviation of features and feed it as an additional feature to the last layer of the network. Both methods use a single layer of handcrafted batch statistics, instead of letting the discriminator learn arbitrary batch statistics useful for discrimination as in our approach. Moreover, in both methods the discriminator still assesses single samples, rather than entire batches. Radford et al. (2015) reported improved results with batch normalization

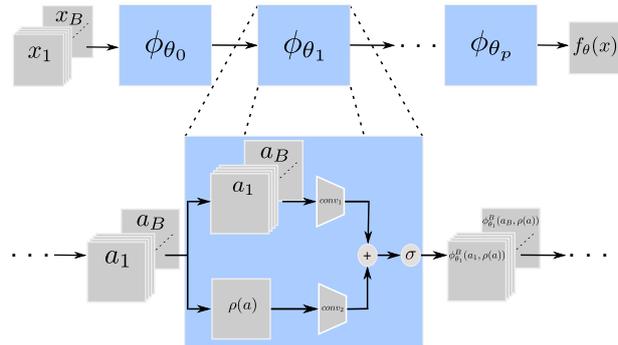

*Figure 1.* Graphical representation of our discriminator architecture. Each convolutional layer of an otherwise classical CNN architecture is modified to include permutation invariant batch statistics, denoted $\rho(x)$. This is repeated at every layer so that the network gradually builds up more complex statistics.

in the discriminator, which may also be due to reliance on batch statistics.

Other works, such as (Li et al., 2015) and (Dziugaite et al., 2015), replace the discriminator with a fixed distributional loss between true and generated samples, the *maximum mean discrepancy*, as the criterion to train the generative model. This has the advantage of relieving the inherent instability of GANs, but lacks the flexibility of an adaptive discriminator.

The discriminator we introduce treats batches as *sets* of samples. Processing sets prescribes the use of permutation invariant networks. There has been a large body of work around permutation invariant networks, e.g (McGregor, 2007; 2008; Qi et al., 2016; Zaheer et al., 2017; Vaswani et al., 2017). Our processing is inspired by (McGregor, 2007; 2008) which designs a special kind of layer that provides the desired invariance property. The network from McGregor (2007) is a multi-layer perceptron in which the single hidden layer performs a batchwise computation that makes the result equivariant by permutation. Here we show that stacking such hidden layers and reducing the final layer with a permutation invariant reduction, covers the whole space of continuous permutation invariant functions.

Zaheer et al. (2017) first process each element of the set independently, then aggregate the resulting representation using a permutation invariant operation, and finally process the permutation invariant quantity. Qi et al. (2016) process 3D point cloud data, and interleave layers that process points independently, and layers that apply equivariant transformations. The output of their networks are either permutation equivariant for pointcloud segmentation, or permutation invariant for shape recognition. In our approach we stack permutation equivariant layers that combine batch information and sample information at every level, and aggregate these in the final layer using a permutation invariant operation.

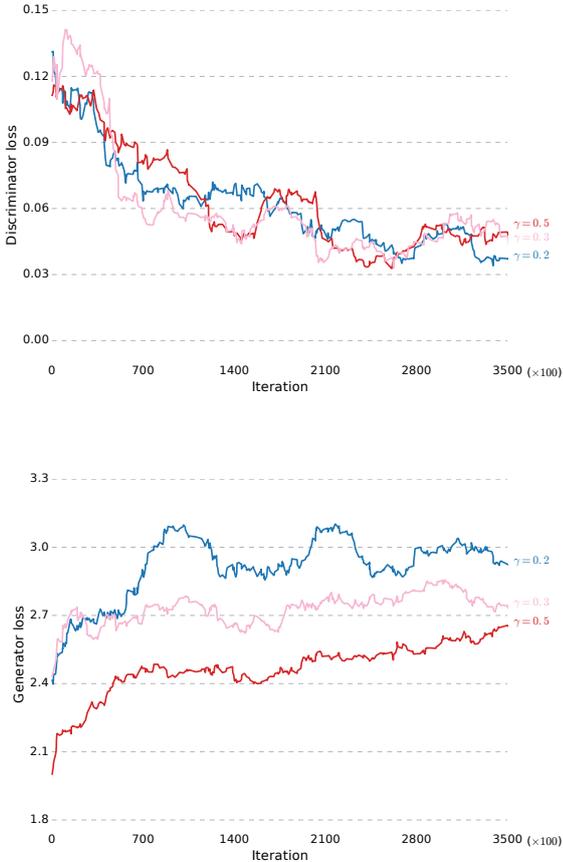

*Figure 2.* Effect of batch smoothing with different $\gamma$'s on the generator and discriminator losses.

More complex approaches to permutation invariance or equivariance appear in (Guttenberg et al., 2016). We prove, however, that our simpler architecture already covers the full space of permutation invariant functions.

Improving the training of GANs has received a lot of recent attention. For instance, Arjovsky et al. (2017), Gulrajani et al. (2017) and Miyato et al. (2018) constrain the Lipschitz constant of the network and show that this stabilizes training and improves performance. Karras et al. (2018) achieved impressive results by gradually increasing the resolution of the generated images as training progresses.

## 3. Adversarial learning with permutation-invariant batch features

Using a batch of samples rather than individual samples as input to the discriminator can provide global statistics about the distributions of interest. Such statistics could be useful to avoid mode dropping. Adversarial learning (Goodfellow et al., 2014) can easily be extended to the batch discrimination case. For a fixed batch size $B$, the corresponding two-player optimization procedure becomes

$$\min_G \max_D \mathbb{E}_{x_1,\ldots,x_B \sim \mathcal{D}} \left[ \log D(x_1, \ldots, x_B) \right] + \quad (1)$$
$$\mathbb{E}_{z_1,\ldots,z_B \sim \mathcal{Z}} \left[ \log(1 - D(G(z_1), \ldots, G(z_B))) \right]$$

with $\mathcal{D}$ the empirical distribution over data, $\mathcal{Z}$ a distribution over the latent variable that is the input of the generator, $G$ a pointwise generator and $D$ a batch discriminator.[1] This leads to a learning procedure similar to the usual GAN algorithm, except that the loss encourages the discriminator to output 1 when faced with an entire batch of real data, and 0 when faced with an entire batch of generated data.

Unfortunately, this basic procedure makes the work of the discriminator too easy. As the discriminator is only faced with batches that consist of either only training samples or only generated samples, it can base its prediction on any subset of these samples. For example, a single poor generated sample would be enough to reject a batch. To cope with this deficiency, we propose to sample batches that mix both training and generated data. The discriminator's task is to predict the *proportion* of real images in the batch, which is clearly a permutation invariant quantity.

### 3.1. Batch smoothing as a regularizer

A naive approach to sampling mixed batches would be, for each batch index, to pick a datapoint from either real or generated images with probability $\frac{1}{2}$. This is necessarily ill behaved: as the batch size increases, the ratio of training data to generated data in the batch tends to $\frac{1}{2}$ by the law of large numbers. Consequently, a discriminator always predicting $\frac{1}{2}$ would achieve very low error with large batch sizes, and provide no training signal to the generator.

Instead, for each batch we sample a ratio $p$ from a distribution $\mathcal{P}$ on $[0, 1]$, and construct a batch by picking real samples with probability $p$ and generated samples with probability $1 - p$. This forces the discriminator to predict across an entire range of possible values of $p$.

Formally, suppose we are given a batch of training data $x \in \mathbb{R}^{B \times n}$ and a batch of generated data $\tilde{x} \in \mathbb{R}^{B \times n}$. To mix $x$ and $\tilde{x}$, a binary vector $\beta$ is sampled from $\mathcal{B}(p)^B$, a $B$-dimensional Bernoulli distribution with parameter $p$. The mixed batch with mixing vector $\beta$ is denoted

$$m_\beta(x, \tilde{x}) := x \odot \beta + \tilde{x} \odot (1 - \beta). \quad (2)$$

---
[1] The generator $G$ could also be modified to produce batches of data, which can help to cover more modes per batch, but this deviates from the objective of learning a density estimator from which we can draw i.i.d. samples.

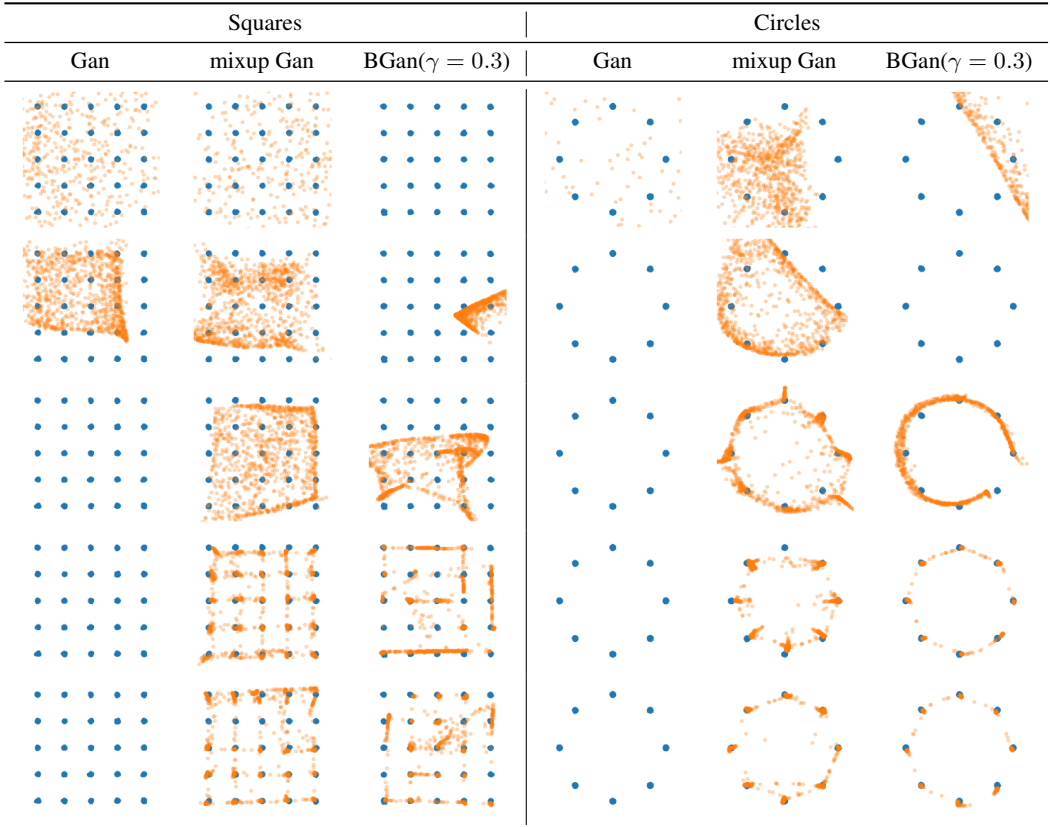

*Figure 3.* Comparison between standard, mixup and batch smoothing GANs on a 2D experiment. Training at iterations 10, 100, 1000, 10000 and 20000.

This apparently wastes some samples, but we can reuse the discarded samples by using $1 - \beta$ in the next batch.

The discriminator has to predict the ratio of real images, $\frac{\#\beta}{B}$ where $\#\beta$ is the sum of the components of $\beta$. As a loss on the predicted ratio, we use the Kullback–Leibler divergence between a Bernoulli distribution with the actual ratio of real images, and a Bernoulli distribution with the predicted ratio. The divergence between Bernoulli distributions with parameters $u$ and $v$ is

$$\mathrm{KL}(\mathcal{B}(u) \,\|\, \mathcal{B}(v)) = u \log \frac{u}{v} + (1-u) \log \frac{1-u}{1-v}. \quad (3)$$

Formally, the discriminator $D$ will minimize the objective

$$\mathbb{E}_{p \sim \mathcal{P},\, \beta \sim \mathcal{B}(p)^B} \mathrm{KL}\left(\mathcal{B}\left(\frac{\#\beta}{B}\right) \,\Big\|\, \mathcal{B}(D(m_\beta(x, \tilde{x})))\right), \quad (4)$$

where the expectation is over sampling $p$ from a distribution $\mathcal{P}$, typically uniform on $[0, 1]$, then sampling a mixed mini-batch. For clarity, we have omitted the expectation over the sampling of training and generated samples

The generator is trained with the loss

$$\mathbb{E}_{p \sim \mathcal{P},\, \beta \sim \mathcal{B}(p)^B} \log(D(m_\beta(x, \tilde{x}))). \quad (5)$$

This loss, which is not the generator loss associated to the min-max optimization problem, is known to saturate less (Goodfellow et al., 2014).

In some experimental cases, using the discriminator loss (4) with $\mathcal{P} = \mathcal{U}([0, 1])$ made discriminator training too difficult. To alleviate some of the difficulty, we sampled the mixing variable $p$ from a reduced symmetric union of intervals $[0, \gamma] \cup [1 - \gamma, 1]$. With low $\gamma$, all generated batches are nearly purely taken from either real or fake data. We refer to this training method as *batch smoothing-$\gamma$*. Batch smoothing-0 corresponds to no mixing, while batch smoothing-0.5 corresponds to equation (4).

### 3.2. The optimal discriminator for batch smoothing

The optimal discriminator for batch smoothing can be computed explicitly, for $p \sim \mathcal{U}([0, 1])$, and extends the usual GAN discriminator when $B = 1$.

**Proposition 1.** *The optimal discriminator for the loss* (4)*,*

*given a batch* $y \in \mathbb{R}^{B \times N}$, *is*

$$D^*(y) = \frac{1}{2} \frac{p_{\text{unbalanced}}(y)}{p_{\text{balanced}}(y)} \qquad (6)$$

*where the distribution* $p_{\text{balanced}}$ *and* $p_{\text{unbalanced}}$ *on batches are defined as*

$$p_{\text{balanced}}(y) = \frac{1}{B+1} \sum_{\beta \in \{0,1\}^B} \frac{p_1(y)^\beta p_2(y)^{1-\beta}}{\binom{B}{\#\beta}}$$

$$p_{\text{unbalanced}}(y) = \frac{2}{B+1} \sum_{\beta \in \{0,1\}^B} \frac{p_1(y)^\beta p_2(y)^{1-\beta}}{\binom{B}{\#\beta}} \frac{\#\beta}{B}. \qquad (7)$$

*in which* $p_1$ *is the data distribution and* $p_2$ *the distribution of generated samples, and where* $p_1(y)^\beta$ *is shorthand for* $p_1(y_1)^{\beta_1} \ldots p_1(y_B)^{\beta_B}$.

The proof is technical and is deferred to the supplementary material. For non-uniform beta distributions on $p$, a similar result holds, with different coefficients depending on $\#\beta$ and $B$ in the sum.

These heavy expressions can be interpreted easily. First, in the case $B = 1$, the optimal discriminator reduces to the optimal discriminator for a standard GAN, $D^* = \frac{p_1(y)}{p_1(y) + p_2(y)}$.

Actually $p_{\text{balanced}}(y)$ is simply the distribution of batches $y$ under our procedure of sampling $p$ uniformly, then sampling $\beta \sim \mathcal{B}(p)^B$. The binomial coefficients put on equal footing contributions with different true/fake ratios.

The generator loss (5), when faced with the optimal discriminator, is the Kullback–Leibler divergence between $p_{\text{balanced}}$ and $p_{\text{unbalanced}}$ (up to sign and a constant $\log(2)$). Since $p_{\text{unbalanced}}$ puts more weight on batches with higher $\#\beta$ (more true samples), this brings fake samples closer to true ones.

Since $p_{\text{balanced}}$ and $p_{\text{unbalanced}}$ differ by a factor $2\#\beta/B$, the ratio $D^* = \frac{1}{2} \frac{p_{\text{unbalanced}}(y)}{p_{\text{balanced}}(y)}$ is simply the expectation of $\#\beta/B$ under a probability distribution on $\beta$ that is proportional to $\frac{p_1(y)^\beta p_2(y)^{1-\beta}}{\binom{B}{\#\beta}}$. But this is the posterior distribution on $\beta$ given the batch $y$ and the uniform prior on the ratio $p$. Thus, the optimal discriminator is just the posterior mean of the ratio of true samples, $D^*(y) = \mathbb{E}_{\beta|y}\left[\frac{\#\beta}{B}\right]$. This is standard when minimizing the expected divergence between Bernoulli distributions and the approach can therefore be extended to non-uniform priors on $p$ as shown in section 9.

## 4. Permutation invariant networks

Computing statistics of probability distributions from batches of i.i.d. samples requires to compute quantities that are invariant to permuting the order of samples within the batch. In this section we propose a permutation equivariant layer that can be used together with a permutation invariant aggregation operation to build networks that are permutation invariant. We also provide a sketch of proof (fully developed in the supplementary material) that this architecture is able to reach all symmetric continuous functions, and only represents such functions.

### 4.1. Building a permutation invariant architecture

A naive way of achieving invariance to batch permutations is to consider the batch dimension as a regular feature dimension, and to randomly reorder the batches at each step. This multiplies the input dimension by the batch size, and thus greatly increases the number of trainable parameters. Moreover, this only provides approximate invariance to batch permutation, as the network has to infer the invariance based on the training data.

Instead, we propose to directly build invariance into the architecture. This method drastically reduces the number of parameters compared to the naive approach, bringing it back in line with ordinary networks, and ensures strict invariance to batch permutation.

Let us first formalize the notion of batch permutation invariance and equivariance. A function $f$ from $\mathbb{R}^{B \times l}$ to $\mathbb{R}^{B \times L}$ is *batch permutation equivariant* if permuting samples in the batch results in the same permutation of the outputs: for any permutation $\sigma$ of the inputs,

$$f(x_{\sigma(1)}, \ldots, x_{\sigma(B)}) = f(x)_{\sigma(1)}, \ldots, f(x)_{\sigma(B)}. \qquad (8)$$

For instance, any regular neural network or other function treating the inputs $x_1, \ldots, x_B$ independently in parallel, is batch permutation equivariant.

A function $f$ from $\mathbb{R}^{B \times l}$ to $\mathbb{R}^L$ is *batch permutation invariant* if permuting the inputs in the batch does not change the output: for any permutation on batch indices $\sigma$,

$$f(x_{\sigma(1)}, \ldots, x_{\sigma(B)}) = f(x_1, \ldots, x_B). \qquad (9)$$

The mean, the max or the standard deviation along the batch axis are all batch permutation invariant.

Permutation equivariant and permutation invariant functions can be obtained by combining ordinary, parallel treatment of batch samples with an additional *batch-averaging* operation that performs an average of the activations across the batch direction. In our architecture, this averaging is the only form of interaction between different elements of the batch. It is one of our main results that such operations are sufficient to recover all invariant functions.

Formally, on a batch of data $x \in \mathbb{R}^{B \times n}$, our proposed batch

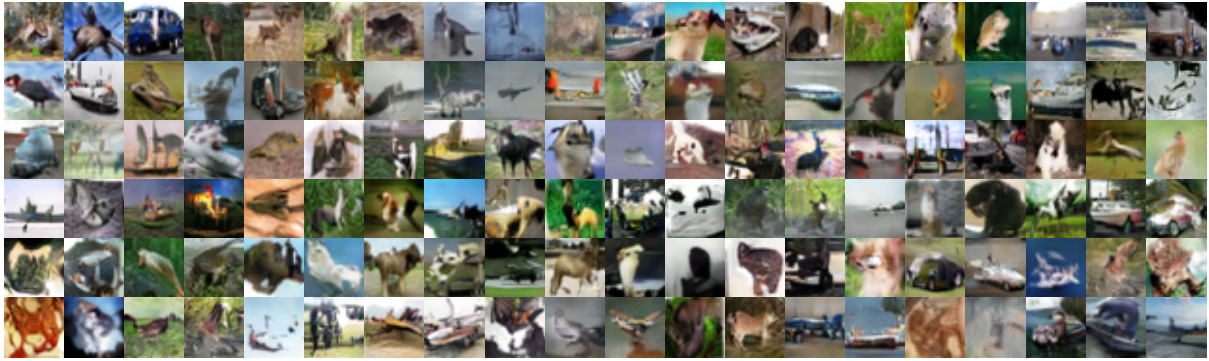

*Figure 4.* Sample images generated by our best model trained on CIFAR10.

permutation invariant network $f_\theta$ is defined as

$$f_\theta(x) = \frac{1}{B} \sum_{b=1}^{B} (\phi_{\theta_p} \circ \phi_{\theta_{p-1}} \circ \ldots \circ \phi_{\theta_0}(x))_b \quad (10)$$

where each $\phi_{\theta_i}$ is a batch permutation equivariant function from $\mathbb{R}^{B \times l_{i-1}}$ to $\mathbb{R}^{B \times l_i}$, where the $l_i$'s are the layer sizes.

The equivariant layer operation $\phi_\theta$ with $l$ input features and $L$ output features comprises an ordinary weight matrix $\Lambda \in \mathbb{R}^{l \times L}$ that treats each data point of the batch independently ("non-batch-mixing"), a batch-mixing weight matrix $\Gamma \in \mathbb{R}^{l \times L}$, and a bias vector $\beta \in \mathbb{R}^L$. As in regular neural networks, $\Lambda$ processes each data point in the batch independently. On the other hand, the weight matrix $\Gamma$ operates after computing an average across the whole batch. Defining $\rho$ as the batch average for each feature,

$$\rho(x_1, \ldots, x_B) := \frac{1}{B} \sum_{b=1}^{B} x_b \quad (11)$$

the permutation-equivariant layer $\phi$ is formally defined as

$$\phi_\theta(x)_b := \mu\Big(\beta + x_b \Lambda + \rho(x) \Gamma\Big) \quad (12)$$

where $\mu$ is a nonlinearity, $b$ is a batch index, and the parameter of the layer is $\theta = (\beta, \Lambda, \Gamma)$.

### 4.2. Networks of equivariant layers provide universal approximation of permutation invariant functions

The networks constructed above are permutation invariant by construction. However, it is unclear a priori that all permutation invariant functions can be represented this way: the functions that can be approximated to arbitrary precision by those networks could be a strict subset of the set of permutation invariant functions. The optimal solution for the discriminator could lie outside this subset, making our construction too restrictive. We now show this is not the case: our architecture satisfies a universal approximation theorem for permutation-invariant functions.

**Theorem 1.** *The set of networks that can be constructed by stacking as in Eq.* (10) *the layers $\phi$ defined in Eq.* (12), *with sigmoid nonlinearities except on the output layer, is dense in the set of permutation-invariant functions (for the topology of uniform convergence on compact sets).*

While the case of one-dimensional features is relatively simple, the multidimensional case is more intricate, and the detailed proof is given in the supplementary material. Let us describe the key ideas underlying the proof.

The standard universal approximation theorem for neural networks proves the following: for any continuous function $f$, we can find a network that given a batch $x = (x_1, \ldots, x_B)$, computes $(f(x_1), \ldots, f(x_B))$. This is insufficient for our purpose as it provides no way of mixing information between samples in the batch.

First, we prove that the set of functions that can be approximated to arbitrary precision by our networks is an algebra, *i.e.*, a vector space stable under products. From this point on, it remains to be shown that this algebra contains a generative family of the continuous symmetric functions.

To prove that we can compute the sum of two functions $f_1$ and $f_2$, compute $f_1$ and $f_2$ on different channels (this is possible even if $f_1$ and $f_2$ require different numbers of layers, by filling in with the identity if necessary). Then sum across channels, which is possible in (12).

To compute products, first compute $f_1$ and $f_2$ on different channels, then apply the universal approximation theorem to turn this into $\log f_1$ and $\log f_2$, then add, then take the exponential thanks to the universal approximation theorem.

The key point is then the following: the algebra of all permutation-invariant polynomials over the components of $(x_1, \ldots, x_B)$ is generated *as an algebra* by the averages $\frac{1}{B}(f(x_1) + \ldots + f(x_B))$ when $f$ ranges over all functions of single batch elements. This non-trivial algebraic statement is proved in the supplementary material.

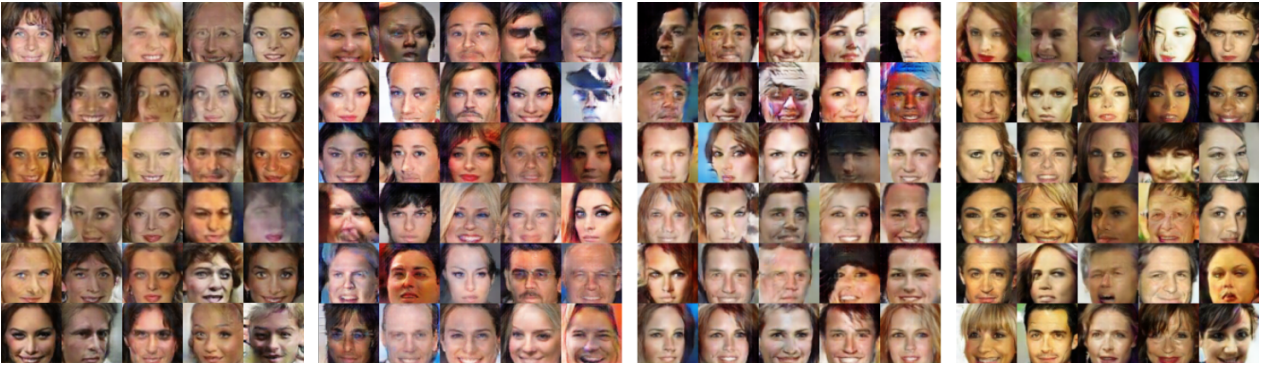

*Figure 5.* Samples obtained after 66000 iterations on the celebA dataset. From left to right: (a) Standard GAN (b) Single batch discriminator, no batch smoothing. (c) Single batch discriminator, batch smoothing $\gamma = 0.5$. (d) Multiple batch discriminators, batch smoothing $\gamma = 0.5$

By construction, such functions $\frac{1}{B}(f(x_1)+\ldots+f(x_B))$ are readily available in our architecture, by computing $f$ as in an ordinary network and then applying the batch-averaging operation $\rho$ in the next layer. Further layers provide sums and products of those thanks to the algebra property. We can conclude with a symmetric version of the Stone–Weierstrass theorem (polynomials are dense in continuous functions).

### 4.3. Practical architecture

In our experiments, we apply the constructions above to standard, deep convolutional neural networks. In practice, for the linear operations $\Lambda$ and $\Gamma$ in (12) we use convolutional kernels (of size $3 \times 3$) acting over $x_b$ and $\rho(x)$ respectively. Weight tensors $\Lambda$ and $\Gamma$ are also reweighted like so that at the start of training $\rho(x)$ does not contribute disproportionately compared with other features: $\tilde{\Lambda} = \frac{|B|}{|B|+1}\Lambda$ and $\tilde{\Gamma} = \frac{1}{|B|+1}\Gamma$ where $|B|$ denotes the size of batch $B$. While these coefficients could be learned, we have found this explicit initialization to improve training. Figure 1 shows how to modify standard CNN architectures to adapt each layer to our method.

In the first setup, which we refer to as BGAN, a permutation invariant reduction is done at the end of the discriminator, yielding a single prediction per batch, which is evaluated with the loss in (4). We also introduce a setup, M-BGAN, where we swap the order of averaging and applying the loss. [2] Namely, letting $y$ be the single target for the batch (in our case, the proportion of real samples), the BGAN case translates into

$$\mathcal{L}((o_1,\ldots,o_B),y) = \ell\left(\frac{1}{B}\sum_{i=1}^{B} o_i, y\right) \quad (13)$$

---
[2]This was initially a bug that worked.

while M-BGAN translates to

$$\mathcal{L}((o_1,\ldots,o_B),y) = \frac{1}{B}\sum_{i=1}^{B} \ell(o_i, y) \quad (14)$$

where $\mathcal{L}$ is the final loss function, $\ell$ is the KL loss function used in (4), $(o_1,\ldots,o_b)$ is the output of the last equivariant layer, and $y$ is the target for the *whole* batch.

Both these losses are permutation invariant. A more detailed explanation of M-BGAN is given in Section 11.

## 5. Experiments

### 5.1. Synthetic 2D distributions

The synthetic dataset from Zhang et al. (2017) is explicitly designed to test mode dropping. The data are sampled from a mixture of concentrated Gaussians in the 2D plane. We compare standard GAN training, "mixup" training (Zhang et al., 2017), and batch smoothing using the BGAN from Section 4.3.

In all cases, the generators and discriminators are three-layer ReLU networks with 512 units per layer. The latent variables of the generator are 2-dimensional standard Gaussians. The models are trained on their respective losses using the Adam (Kingma & Ba, 2015) optimizer, with default parameters. The discriminator is trained for five steps for each generator step.

The results are summarized in Figure 3. Batch smoothing and mixup have similar effects. Results for BGAN and M-BGAN are qualitatively similar on this dataset and we only display results for BGAN. The standard GAN setting quickly diverges, due to its inability to fit several modes simultaneously, while both batch smoothing and mixup successfully fit the majority of modes of the distribution.

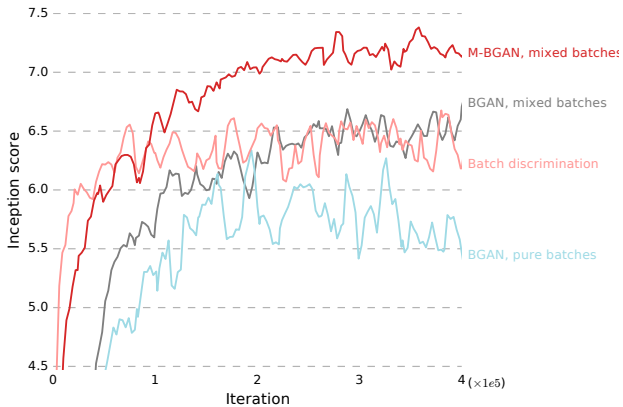

*Figure 6.* Inception score for various versions of BGAN and for batch discrimination (Salimans et al., 2016).

### 5.2. Experimental results on CIFAR10

Next, we consider image generation on the CIFAR10 dataset. We use the simple architecture from (Miyato et al., 2018), minimally modified to obtain permutation invariance thanks to (12). All other architectural choices are unchanged. The same Adam hyperparameters from (Miyato et al., 2018) are used for all models: $\alpha = 2e^{-4}$, $\beta_1 = 0.5$, $\beta_2 = 0.999$, and no learning rate decay. We performed hyperparameter search for the number of discrimination steps between each generation step, $n_{\text{disc}}$, over the range $\{1, \ldots, 5\}$, and for the batch smoothing parameter $\gamma$ over $[0.2, 0.5]$. All models are trained for $400,000$ iterations, counting both generation and discrimination steps. We compare smoothed BGAN and M-BGAN, and the same network trained with spectral normalization (Miyato et al., 2018) (SN), and gradient penalty (Gulrajani et al., 2017) on both the Wasserstein (Arjovsky et al., 2017) (WGP) and the standard loss (GP). We also compare to a model using the batch-discrimination layer from (Salimans et al., 2016), adding a final batch discrimination layer to the architecture of (Miyato et al., 2018). All models are evaluated by reporting the Inception Score and the Fréchet Inception Distance (Heusel et al., 2017) and results are summarized in Table 2. Figure 4 displays sample images generated with our best model.

Figure 5.2 highlights the training dynamics of each model[3]. On this architecture, M-BGAN heavily outperforms both batch discrimination and our other variants, and yields results similar to, or slightly better than (Miyato et al., 2018). Model trained with batch smoothing display results on par with batch discrimination, and much better than without batch smoothing.

[3] For readability, a slight smoothing is performed on the curves.

*Table 1.* Comparison to the state of the art in terms of inception score (IS) and Fréchet inception distance (FID).

| Model | IS | FID |
|---|---|---|
| WGP (Miyato et al., 2018) | $6.68 \pm .06$ | 40.2 |
| GP (Miyato et al., 2018) | $6.93 \pm .08$ | 37.7 |
| SN (Miyato et al., 2018) | $7.42 \pm .08$ | 29.3 |
| Salimans et al. | $7.09 \pm .08$ | 35.0 |
| BGAN | $7.05 \pm .06$ | 36.47 |
| M-BGAN | $7.49 \pm .06$ | 23.71 |

### 5.3. Effect of batch smoothing on the generator and discriminator losses

To check the effect of the batch smoothing parameter $\gamma$ on the loss, we plot the discriminator and generator losses of the network for different $\gamma$'s. The smaller the $\gamma$, the purer the batches. We would expect discriminator training to be more difficult with larger $\gamma$. The results corroborate this insight (Fig. 2). BGAN and M-BGAN behave similarly and we only report on BGAN in the figure. The discriminator loss is not directly affected by an increase in $\gamma$, but the generator loss is lower for larger $\gamma$, revealing the relative advantage of the generator on the discriminator.

This suggests to increase $\gamma$ if the discriminator dominates learning, and to decrease $\gamma$ if the discriminator is stuck at a high value in spite of poor generated samples.

### 5.4. Qualitative results on celebA

Finally, on the celebA face dataset, we adapt the simple architecture of (Miyato et al., 2018) to the increased resolution by adding a layer to both networks. For optimization we use Adam with $\beta_1 = 0, \beta_2 = 0.9, \alpha = 1e-4$, and $n_{\text{disc}} = 1$. Fig. 5 dislays BGAN samples with pure batches, and BGAN and M-BGAN samples with $\gamma = .5$. The visual quality of the samples is reasonable; we believe that an improvement is visible from pure batches to M-BGAN.

## 6. Conclusion

We introduced a method to feed batches of samples to the discriminator of a GAN in an principled way, based on two observations: feeding all-fake or all-genuine batches to a discriminator makes its task too easy; second, a simple architectural trick makes it possible to provably recover all functions of the batch as an unordered set. Experimentally, this provides a new, alternative method to reduce mode dropping and reach good quantitative scores in GAN training.

# 7. Supplementary material

In what follows, we aim at proving a universal approximation theorem for the class of permutation invariant neural networks we have defined. To ease readings, products, sums and real function applications are assumed to be broadcasted when need be. Throughout the paper the batch dimension $n$ is constant and ommited from set indices.

**Definition 1.** *A function $f\colon \mathbb{R}^{n\times k} \mapsto \mathbb{R}^l$ is* symmetric *if for any permutation of indexes $\sigma$ and for all $x \in \mathbb{R}^{n\times k}$, $f(x_{\sigma(1)}, \ldots, x_{\sigma(n)}) = f(x_1, \ldots, x_n)$. The set of continuous symmetric functions from $\mathbb{R}^{n\times k}$ to $\mathbb{R}^l$ is denoted by $\mathcal{I}_k^l$*

**Definition 2.** *A function $f\colon \mathbb{R}^{n\times k} \mapsto \mathbb{R}^{n\times l}$ is* permutation equivariant *if for any permutation of indexes $\sigma$ and for al $x \in \mathbb{R}^n$, $f(x_{\sigma(1)}, \ldots, x_{\sigma(n)}) = f(x)_{\sigma(1)}, \ldots, f(x)_{\sigma(n)}$.*

When symmetric functions and permutation equivariant functions are restricted to a compact, we assume that the compact itself is symmetric.

In what follows, we use $\rho$ as a reducing operator on vectors defined for $x \in \mathbb{R}^{n\times k}$ by

$$\rho(x)_j = \frac{1}{n}\sum_{i=1}^n x_{i,j}. \qquad (15)$$

**Definition 3.** *Let the sets $E_k^l$ be sets that contain* permutation equivariant neural networks *from $\mathbb{R}^{n\times k}$ to $\mathbb{R}^{n\times l}$, recursively defined thus:*

- *For all $k \in \mathbb{N}$, the identity function on $\mathbb{R}^{n\times k}$ belongs to $E_k^k$.*
- *For all $f \in E_r^k$, $\Gamma \in \mathbb{R}^{l\times k}$, $\Lambda \in \mathbb{R}^{l\times k}$ and $\beta \in \mathbb{R}^l$, and for act, a sigmoid activation function, $g$ defined as*

$$g(x)_{i,j} = \sum_{p=1}^k \Gamma_{j,p} act(f(x))_{i,p} + \sum_{p=1}^k \Lambda_{j,p} \rho(act \circ f(x))_p + \beta_j) \qquad (16)$$

*is in $E_r^l$.*

*The number of layers of the network is defined as the induction depth of the previous construction. The set of thus constructed* permutation equivariant neural networks *with number of layers $L$ is denoted by $E(L)_k^l$. Note that this class of function is trivially stable by composition, i.e. if $g_1 \in E_{l_1}^{l_2}$ and $g_2 \in E_{l_2}^{l_3}$, the $g_2 \circ g_1 \in E_{l_1}^{l_3}$.*

**Definition 4.** *Let $I_k^l$ be a set containing* symmetric neural networks *from $\mathbb{R}^{n\times k}$ to $\mathbb{R}^l$ defined as*

$$I_k^l = \rho(E_k^l). \qquad (17)$$

We have constructed sets $I_k^l$, containing permutation invarient networks. We now show that the way we they are constructed is not too restrictive, i.e. that any analytical symmetric function can be approximated with arbitrary precision by a sufficiently expressive network of our construct. In other words we aim at proving Theorem 1 2.

**Theorem 2.** *For all $n, k, l$ and for all compact $K$, $I_k^l\big|_K$ is dense in $\mathcal{I}_k^l\big|_K$.*

The first step of the proof is to show that the closure of $I_k^l\big|_K$ is a ring, i.e. that it is stable by sum, product and that each element has an inverse for $+$, as well as a vectorial space, making it an algebra. The second step is to prove that this closure contains a generative family of the set of all polynomials that operate symmetrically on the batch dimension and because symmetric polynomials are dense in the set of all symmetric functions, this proves the theorem.

**Lemma 1.** *If $f_1 \in \overline{E_{l_1}^{l_2}\big|_K}$ and $f_2 \in \overline{E_{l_2}^{l_3}\big|_{f_1(K)}}$ then $f_2 \circ f_1 \in \overline{E_{l_1}^{l_3}\big|_K}$.*

*Proof.* Let $\varepsilon > 0$, $f_2$ is continuous on a compact set, thus uniformly continuous, and there exists an $\eta > 0$ such that $\|x - x'\| < \eta$ implies $\|f_2(x) - f_2(x')\| < \frac{\varepsilon}{2}$. Now let $g_1 \in E_{l_1}^{l_2}\big|_K$ be such that $\|g_1 - f_1\|_\infty \le \eta$ and $g_2 \in E_{l_2}^{l_3}\big|_K$ such that $\|g_2 - f_2\|_\infty \le \frac{\varepsilon}{2}$, then, for $x$ in $K$

$$\|f_2 \circ f_1(x) - g_2 \circ g_1(x)\| \le \|f_2 \circ f_1(x) - g_2 \circ f_1(x)\| + \|g_2 \circ f_1(x) - g_2 \circ g_1(x)\|$$
$$\le \varepsilon$$

□

**Lemma 2.** *For any continuous functions $g\colon \mathbb{R}^k \mapsto \mathbb{R}^l$, the restriction of the function $G\colon \mathbb{R}^{n\times k} \mapsto \mathbb{R}^{n\times k}$, defined as $G(x) = (g(x_1), \ldots, g(x_n))$, to a compact $K$ is in $\overline{E_k^l\big|_K}$. More precisely, for all $L \geq 2$, the restriction of $G$ to $K$ is in $\overline{E(L)_k^l\big|_K}$.*

*Proof.* This is a consequence of the neural network universal approximation theorem, as stated e.g. in (Cybenko, 1989). □

**Lemma 3.** *If $f_1 \in E_k^{l_1}\big|_K$, $f_2 \in E_k^{l_2}\big|_K$ and $f_1$ and $f_2$ have the same number of layers (i.e. they have the same induction depth), then $\mathrm{concat}_1(f_1, f_2) \in E_k^{l_1, l_2}\big|_K$, with*

$$\mathrm{concat}_1(x,y)_{i,j} = \begin{cases} x_{i,j} \text{ if } j \leq l_1 \\ y_{i,j-l_1} \text{ otherwise} \end{cases} \tag{18}$$

*Proof.* By induction on the number of layers $L$,

- if $L = 0$, the result is clear.
- if $L > 0$, let $g_1$, $\Gamma_1$, $\Lambda_1$ and $\beta_1$ as well as $g_2$, $\Gamma_2$, $\Lambda_2$ and $\beta_2$ be the parameters associated to $f_1$ and $f_2$, then, by induction, $\mathrm{concat}_1(g_1, g_2)$ is a permutation equivariant network, and $\mathrm{concat}_1(f_1, f_2)$ is obtained by setting $\Gamma$ to be the block diagonal matrix obtained with $\Gamma_1$ and $\Gamma_2$, $\Lambda$, the block diagonal matrix obtained with $\Lambda_1$ and $\Lambda_2$, and $\beta$ the concatenation of both $\beta$'s.

□

**Lemma 4.** *If $f_1 \in \overline{E_k^{l_1}\big|_K}$, $f_2 \in \overline{E_k^{l_2}\big|_K}$, then $\mathrm{concat}_1(f_1, f_2) \in \overline{E_k^{l_1+l_2}\big|_K}$.*

*Proof.* Let $\varepsilon > 0$, let $g_1 \in E_k^{l_1}\big|_K$ and $g_2 \in E_k^{l_2}\big|_K$ be such that $\|g_1 - f_1\|_\infty \leq \frac{\varepsilon}{4}$ and $\|g_2 - f_2\|_\infty \leq \frac{\varepsilon}{4}$. Denote by $L_1$ and $L_2$ the numbers of layers of $g_1$ and $g_2$. We assume $L_1 \geq L_2$ without loss of generality. By lemma 2, there exist $h_1 \in E_{l_1}^{l_1}\big|_K$ and $h_2 \in E_{l_2}^{l_2}\big|_K$ with $h_1$ of depth 2 and $h_2$ of depth $L_1 - L_2 + 2$ such that $\|h_1 - Id\|_\infty \leq \frac{\varepsilon}{4}$ on $g_1(K)$ and $\|h_2 - Id\|_\infty \leq \frac{\varepsilon}{4}$ on $g_2(K)$. The networks $h_1 \circ g_1$ and $h_2 \circ g_2$ have the same number of layers, consequently, $\mathrm{concat}_1(h_1 \circ g_1, h_2 \circ g_2) \in E_k^{l_1, l_2}\big|_K$. Besides,

$$\|\mathrm{concat}_1(f_1, f_2) - \mathrm{concat}_1(h_1 \circ g_1, h_2 \circ g_2)\|_\infty \tag{19}$$
$$\leq \|f_1 - g_1\|_\infty + \|h_1 \circ g_1 - g_1\|_\infty + \|f_2 - g_2\|_\infty + \|h_2 \circ g_2 - g_2\|_\infty \tag{20}$$
$$\leq \varepsilon \tag{21}$$

yielding the result. □

**Lemma 5.** *If $f_1$ and $f_2$ are in $\overline{E_k^l\big|_K}$, then $f_1 + f_2$ is too.*

*Proof.* By lemma 3, $\mathrm{concat}_1(f_1, f_2)$ is in $\overline{E_k^{2l}\big|_K}$. Consider the layer $g$, with kernels $\Gamma_{i,j} = \begin{cases} 1 \text{ if } j = i \text{ or } j = k+i \\ 0 \text{ otherwise} \end{cases}$, $1 \leq i \leq l, 1 \leq j \leq 2l$, $\Lambda = 0$, $\beta = 0$. By lemma 1, as both $\mathrm{concat}_1(f_1, f_2)$ and $g$ are in closures of permutation equivariant networks, their composition is too. This composition is $\mathrm{act}(f_1 + f_2)$. By the universal approximation theorem $\mathrm{act}^{-1}$ is also in the closure so $f_1 + f_2$ is in the closure. □

More generally, following similar reasonings, closures of permutation equivariant networks are vectorial spaces. It follows that closures of permutation invariant networks are vectorial spaces too.

**Lemma 6.** *If $f \in \overline{I_k^l|_K}$, then $F$ defined by*

$$F(x)_{i,j} = f(x)_j \tag{22}$$

*for all $i, j$, is in $\overline{E_k^l|_K}$*

*Proof.* By definition, for any $\varepsilon > 0$, there exists a $G$ in $E_k^l|_K$ such that $f$ and $\rho(G)$ are at distance at most $\frac{\varepsilon}{2}$. Let $\alpha$ be a non zero real number such that $\text{act}^{-1}(\alpha G(x))$ is well defined for any $x \in K$. Consider the equivariant layer

$$m(x)_{i,j} = \alpha^{-1} \rho(\text{act}(x))_j. \tag{23}$$

Let $\eta_1$ be a positive real number, and $L_{\eta_1}$ be a compact set that contains both $\text{act}^{-1}(\alpha G(K))$ and any ball of radius $\eta_1$ contained in this set. $m$ is uniformly continuous on $L_{\eta_1}$, and consequently there exists an $\eta_2$ such that if $x$ and $y$ are at distance at most $\eta_2$, $m(x)$ and $m(y)$ are at distance at most $\frac{\varepsilon}{2}$. Now, by composition and the universal approximation theorem, let $h \in E_k^l$ be such that $h$ and $\text{act}^{-1}(\alpha G)$ are at distance at most $\min(\eta_1, \eta_2)$. Then $m \circ \text{act}^{-1}(\alpha G)$ and $m \circ h$ are at distance at most $\frac{\varepsilon}{2}$, and by triangular inequality, $F$ and $m \circ h$ are at distance at most $\varepsilon$. □

**Lemma 7.** *If $f_1$ and $f_2$ are in $\overline{I_k^l|_K}$, then $f_1 f_2$ is too.*

*Proof.* Let $F_1$ and $F_2$ be the extensions of $f_1$, $f_2$ as defined in lemma 6. There exists a $C \in \mathbb{R}$ such that for all $i, j, x \in K$, $F_1(x)_{i,j} + C > 0$, and similarly for $F_2$. Consequently, by lemma 1, lemma 2 and lemma 5, $\exp(\log(F_1 + C) + \log(F_2 + C)) = F_1 F_2 + F_1 C + F_2 C + C^2 \in \overline{E_k^l|_K}$. As this closure is a vectorial space, $F_1 F_2 \in \overline{E_k^l|_K}$. Consequently, $f_1 f_2 = \rho(F_1 F_2) \in \overline{I_k^l|_K}$. □

We have now shown that $\overline{I_k^l|_K}$ is a ring. We are left to prove that it contains a generative family of the continuous symmetric functions. Let us first exhibit a familly of continuous symmetric functions that is contained in the set of interest, and that we will later show generate all continuous symmetric function.

**Lemma 8.** *For all $f$, restriction of a function from $R^l$ to $R^k$ to a compact set $K$, the symmetric function $F$, defined on $K^{n \times l}$ by*

$$F(x) = \sum_{i=1}^n f(x_i) \tag{24}$$

*is in $I_k^l$.*

*Proof.* By the universal approximation theorem, $f$ is in $\overline{I_k^l|_K}$. By lemma 6, there exists a $G$ in $\overline{E_k^l|_K}$ that replicates $f$ along the batch axis of an equivariant network. Consequently, $\rho(G) = F$ is in $\overline{I_k^l|_K}$. □

We are going to prove that this family of functions generates the set of all symmetric polynomials. Deriving a generalization of Stone Weierstrass theorem to symmetric functions, we obtain the final result.

To keep things general, in what follows, $X$ denotes an arbitrary set, $F$ an algebra of functions on $X$, and $S$ is the symmetrization operator on functions of $X^n$, i.e. for all $(x_1, \ldots, x_n) \in X^n$,

$$(Sf)(x_1, \ldots, x_n) = \sum_\sigma f(x_{\sigma(1)}, \ldots, x_{\sigma(n)}) \tag{25}$$

where the sum is over all permutations of $[1, n]$.

Let $P$ be the algebra of functions of $X^n$ generated by the functions $f(x_k) \colon x \to f(x_k)$ for $f$ in $F$, with a slight abuse of notations. $P$ is linearly generated by the monomials $f_1(x_1) \ldots f_n(x_n)$ for $f_k$ arbitrary functions of $F$. We are interested in the symmetrization of $P$, $SP$. By linearity of $S$, $SP$ is generated by the symmetrized monomials,

$$S f_1(x_1) \ldots f_n(x_n) = \sum_\sigma \prod_{k=1}^n f_k(x_{\sigma(k)}). \tag{26}$$

**Lemma 9.** *SP is generated as an algebra by $Sf(x_1)$ for $f \in F$. Notably, $Sf(x_1)$ takes the special form*

$$Sf(x_1) = \sum_{\sigma} f(x_{\sigma(1)}) = (n-1)! \sum_{k=1}^{n} f(x_k). \tag{27}$$

Typically, for our case, $X = \mathbb{R}^l$ for $l$ the number of input features, $F$ is an algebra of functions containing the multivariate polynomials on $\mathbb{R}^l$, and $SP$ thus contains the set of all polynomials which are symmetric along the batch dimension.

*Proof.* Call *rank* of a monomial $f_1(x_1) \ldots f_n(x_n)$, the number of functions $f_k$ such that $f_k \neq 1$. Let $k_1, \ldots, k_r$ be these indices. Up to renaming $f_{k_1}$ to $f_1$, etc., the monomial can be written as $f_1(x_{k_1}) \ldots f_r(x_{k_r})$.

We will work by induction on $r$. For $r = 1$ the claim is trivial.

Since $S$ does not care about permuting the variables, we have

$$Sf_1(x_{k_1}) \ldots f_r(x_{k_r}) = Sf_1(x_1) \ldots f_r(x_r) = \sum_{\sigma \in \mathcal{S}_K} \prod_{i=1}^{r} f_i(x_{\sigma(i)}) \tag{28}$$

and now the values $\sigma(r+1), \ldots, \sigma(n)$ have no influence so that

$$Sf_1(x_1) \ldots f_n(x_n) = (n-r)! \sum_{\sigma \in \mathrm{Inj}_r^n} \prod_{i=1}^{r} f_i(x_{\sigma(i)}) \tag{29}$$

where $\mathrm{Inj}_r^n$ is the set of injective functions from $r$ to $n$.

Assume we can generate all symmetric monomials up to rank $r$. By definition we can generate $Sf_{r+1}(x_1)$ for any $f_{r+1} \in F$. Then we can generate the product

$$\frac{1}{(n-r-1)!}(Sf_{r+1}(x_1)) \left( \sum_{\sigma \in \mathrm{Inj}_r^n} \prod_{i=1}^{r} f_i(x_{\sigma(i)}) \right) = \left( \sum_{k \in n} f_{r+1}(x_k) \right) \left( \sum_{\sigma \in \mathrm{Inj}_r^n} \prod_{i=1}^{r} f_i(x_{\sigma(i)}) \right)$$

$$= \sum_{\sigma \in \mathrm{Inj}_r^n} \sum_{k \in n} f_{r+1}(x_k) \prod_{i=1}^{r} f_i(x_{\sigma(i)})$$

Now, for each $\sigma$, we can decompose according to whether $k \in \mathrm{Im}\,\sigma$ or $k \in n \setminus \mathrm{Im}\,\sigma$, where $\mathrm{Im}\,\sigma = \{\sigma(1), \ldots, \sigma(r)\}$ is the image of $\sigma$. We obtain two terms

$$\ldots = \sum_{\sigma \in \mathrm{Inj}_r^n} \sum_{k \in \mathrm{Im}\,\sigma} f_{r+1}(x_k) \prod_{i=1}^{r} f_i(x_{\sigma(i)}) + \sum_{\sigma \in \mathrm{Inj}_r^n} \sum_{k \in n \setminus \mathrm{Im}\,\sigma} f_{r+1}(x_k) \prod_{i=1}^{r} f_i(x_{\sigma(i)})$$

But if $k$ is not in $\mathrm{Im}\,\sigma$, then $(\sigma(1), \ldots, \sigma(r), k)$ is an injective function from $r+1$ to $n$. So summing over $\sigma$ then on $k \in n \setminus \mathrm{Im}\,\sigma$ is exactly equivalent to summing over $\sigma \in \mathrm{Inj}_{r+1}^n$. So the second term above is

$$\sum_{\sigma \in \mathrm{Inj}_{r+1}^n} \left( \prod_{i=1}^{r} f_i(x_{\sigma(i)}) \right) f_{r+1}(\sigma(r+1)) = \sum_{\sigma \in \mathrm{Inj}_{r+1}^n} \prod_{i=1}^{r+1} f_i(x_{\sigma(i)}) = Sf_1(x_{k_1}) \ldots f_{r+1}(x_{k_{r+1}})$$

which is the one we are interested in.

So if we prove that we can generate the first term, we are done.

Let us consider the first term, with $k \in \mathrm{Im}\,\sigma$. Now, since $k \in \mathrm{Im}\,\sigma$, we can decompose over the cases $k = \sigma(1), \ldots, k = \sigma(r)$, namely,

$$\sum_{\sigma \in \mathrm{Inj}_r^n} \sum_{k \in \mathrm{Im}\,\sigma} f_{r+1}(x_k) \prod_{i=1}^{r} f_i(x_{\sigma(i)}) = \sum_{\sigma \in \mathrm{Inj}_r^n} \sum_{j=1}^{r} f_{r+1}(x_{\sigma(j)}) \prod_{i=1}^{r} f_i(x_{\sigma(i)}) \tag{30}$$

$$= \sum_{j=1}^{r} \sum_{\sigma \in \mathrm{Inj}_r^n} \prod_{i=1}^{r} \tilde{f}_{ij}(x_{\sigma(i)}) \tag{31}$$

where

$$\tilde{f}_{ij} := \begin{cases} f_i & i \neq j \\ f_i f_{r+1} & i = j \end{cases} \quad (32)$$

Now since $F$ is a ring, $f_i f_{r+1} \in F$. For each $j$ the term

$$\sum_{\sigma \in \text{Inj}_r^n} \prod_{i=1}^r \tilde{f}_{ij}(x_{\sigma(i)}) \quad (33)$$

is equal to $S\tilde{f}_{1j} \ldots \tilde{f}_{rj}$ up to a factor $(n-(r+1))!$. By our induction hypothesis, each term can be generated. This ends the proof. □

**Lemma 10.** *For any compact $K$, any $l \in \mathbb{N}$, the intersection of $\mathcal{I}_l^1$ with the set of multivariate polynomials is dense in $\mathcal{I}_l^1$ for the infinity norm.*

*Proof.* Let $\varepsilon > 0$, and $f$ be in $I_l^1$. There exists a multivariate polynomials $P$ such that $\|P - f\|_\infty \leq \varepsilon$. Let us consider the symmetrized polynomial

$$\tilde{P}(x_1, \ldots, x_n) = \frac{1}{n!} \sum_\sigma P(x_{\sigma(1)}, \ldots, x_{\sigma(n)}). \quad (34)$$

Then $\tilde{P}$ is in the intersection, and, for $x \in K$,

$$\|\tilde{P}(x) - f(x)\| = \|\frac{1}{n!} \sum_\sigma (P(x_{\sigma(1)}, \ldots, x_{\sigma(n)}) - f(x_{\sigma(1)}, \ldots, x_{\sigma(n)}))\| \quad (35)$$

$$\leq \frac{1}{n!} \sum_\sigma \|P(x_{\sigma(1)}, \ldots, x_{\sigma(n)}) - f(x_{\sigma(1)}, \ldots, x_{\sigma(n)})\| \quad (36)$$

$$\leq \varepsilon. \quad (37)$$

□

We now have all the ingredients to end the proof. For a given compact $K$ of $\mathbb{R}^l$, for any multivariate polynomial $P$ of $\mathbb{R}^l$, any $\varepsilon > 0$, there trivially exists an element $f$ of $I_k^1$ at distance at most $\varepsilon$ of $x \to \sum_{i=1}^n P(x_i)$. This means that the closure of the considered set contains all such functions. As this closure is an algebra (it is both a ring and a vectorial space), by lemma 8, it contains the intersection of $\mathcal{I}_l^2$ with the set of multivariate polynomials. By lemma 10, it contains $\mathcal{I}_l^1$, which ends the proof.

## 8. Other details

### 8.1. $p_{balanced}$ and $p_{unbalanced}$ are well normalized:

We now show that $p_{unbalanced}$ is well defined. The computation for $p_{balanced}$ is almost identical and left to the reader.

$$\begin{aligned}
\int_y p_{unbalanced}(y) dy &= \frac{2}{B+1} \sum_{\beta \in \{0,1\}^B} \frac{\#\beta}{BC_B^{\#\beta}} \int_y p_x(y)^\beta p_{\tilde{x}}(y)^{1-\beta} dy \\
&= \frac{2}{B+1} \sum_{\#\beta=1}^B C_B^{\#\beta} \frac{\#\beta}{BC_B^{\#\beta}} \\
&= \frac{2}{(B+1)B} \sum_{\#\beta=1}^B \#\beta \\
&= \frac{2}{(B+1)B} \frac{B(B+1)}{2} \\
&= 1
\end{aligned}$$

# 9. Optimal discriminator for general beta prior

We hereby give a derivation of the optimal discriminator expression, when mixing parameters, $p$'s are drawn from Beta$(a,b)$. This extends Eq. (7), as Beta$(1,1) = \mathcal{U}([0,1])$.

**Beta prior on batch mixing proportion.** Consider mixed batches of samples of size $B$. The $i$-th sample of the batch is a real sample if $\beta_i = 1$ and a false sample if $\beta_i = 0$. Given a certain mixing proportion $p$, assuming that sample origine are sampled independantly according to a Bernoulli of parameter $p$, the probability of a certain $\beta$ is

$$\mathbb{P}(\beta \mid p) = \prod_i p^{\beta_i}(1-p)^{1-\beta_i}, \tag{38}$$

Considering a beta prior distribution Beta$(a,b)$ on the mixing parameter $p \in [0,1]$, the posterior distribution on the number of real sample in the batch $\#\beta = \sum_i \beta_i$ is given by the beta-binomial compound distribution

$$\mathbb{P}(\#\beta) = \int_p \text{Beta}(p \mid a,b) \mathbb{P}(\#\beta \mid p) \tag{39}$$

$$= \binom{B}{\#\beta} \frac{\mathcal{B}(\#\beta + a, B - \#\beta + b)}{\mathcal{B}(a,b)} \tag{40}$$

where $\mathcal{B}(\cdot,\cdot)$ is the beta function. For $a=1, b=1$, i.e. a uniform distribution on mixing parameters, the beta-binomial compound distribution reduces to a uniform distribution on $\#\beta$. From the expression of $\mathbb{P}(\#\beta)$ it follows that

$$\mathbb{P}(\beta) = \frac{\mathcal{B}(\#\beta + a, B - \#\beta + b)}{\mathcal{B}(a,b)}. \tag{41}$$

**Optimal discriminator.** Let $y = m_\beta(x, \tilde{x})$ denote a mixed batch of samples. The discriminator minimizes the KL divergence between $D(y)$ and $\beta$, averaged over batches and mixing vectors $\beta$, see Eq. (4) in the main paper. This reduces to minimizing the expected cross-entropy. For a given batch and mixing vector $\beta$,

$$L(D(y), \#\beta) = -\frac{\#\beta}{B} \ln D(y) - \frac{B - \#\beta}{B} \ln(1 - D(y)). \tag{42}$$

Averaging over batches and mixing vectors,

$$\mathbb{E}_{\beta,y}[L(D(y), \#\beta)] = \int_y \mathbb{P}(y) \sum_\beta \mathbb{P}(\beta \mid y) L(D(y), \#\beta) \tag{43}$$

$$= -\int_y \mathbb{P}(y) \left[ \mathbb{E}_{\beta \mid y}\left[\frac{\#\beta}{B}\right] \ln D(y) + \mathbb{E}_{\beta \mid y}\left[\frac{B-\#\beta}{B}\right] \ln(1 - D(y)) \right] \tag{44}$$

From the latter it yields that for any $y$, the optimal discriminator value $D^*(y)$ is

$$D^*(y) = \mathbb{E}_{\beta \mid y}\left[\frac{\#\beta}{B}\right], \tag{45}$$

*i.e.* the posterior expectation of the fraction of training samples in the batch.

**Posterior analysis.** Through Bayes rule, the posterior expectation yields

$$D^*(y) = \mathbb{E}_{\beta \mid y}\left[\frac{\#\beta}{B}\right] = \frac{\sum_\beta \frac{\#\beta}{B} \mathbb{P}(y \mid \beta) \mathbb{P}(\beta)}{\mathbb{P}(y)}. \tag{46}$$

The marginal on the batch $y$ is

$$\mathbb{P}(y) = \sum_\beta \mathbb{P}(y \mid \beta) \mathbb{P}(\beta) \tag{47}$$

$$= \sum_\beta \mathbb{P}(y \mid \beta) \frac{\mathcal{B}(\#\beta + a, B - \#\beta + b)}{\mathcal{B}(a,b)}. \tag{48}$$

The numerator in Eq. (46) can be written as a distribution on $y$,

$$\mathbb{Q}(y) = \sum_\beta \mathbb{P}(y|\beta)\mathbb{Q}(\beta) \qquad (49)$$

$$\mathbb{Q}(\beta) = \frac{a+b}{a}\mathbb{P}(\beta)\frac{\#\beta}{B}. \qquad (50)$$

The distribution $\mathbb{Q}$ on $\beta$ sums to 1, as $\mathbb{E}_{\mathbb{P}(\#\beta)}[\#\beta] = \frac{Ba}{a+b}$.

This finally yields

$$D^*(y) = \frac{a}{a+b}\frac{\mathbb{Q}(y)}{\mathbb{P}(y)}, \qquad (51)$$

which for the uniform prior on $p$ simplifies to

$$D^*(y) = \frac{1}{2}\frac{\mathbb{Q}(y)}{\mathbb{P}(y)}. \qquad (52)$$

**Expressing $\mathbb{P}(y \mid \beta)$.** Notice that $m_\beta(x, \tilde{x}) = y$ is equivalent to for all i in $\{1, ..., B\}$, $x_i = y_i$ and $\beta_i = 1$ or $\tilde{x}_i = y_i$ and $\beta_i = 0$. Denote by $p_1$ (resp. $p_2$) the distribution of real samples (resp. generated samples).

From the previous observation, it yields that

$$\mathbb{P}(y \mid \beta) = \prod_{i=1}^{B} p_1(y_i)^{\beta_i} p_2(y_i)^{1-\beta_i}. \qquad (53)$$

From the latter and Eq. (52) we obtain the optimal discriminator expression.

## 10. Additional experiments

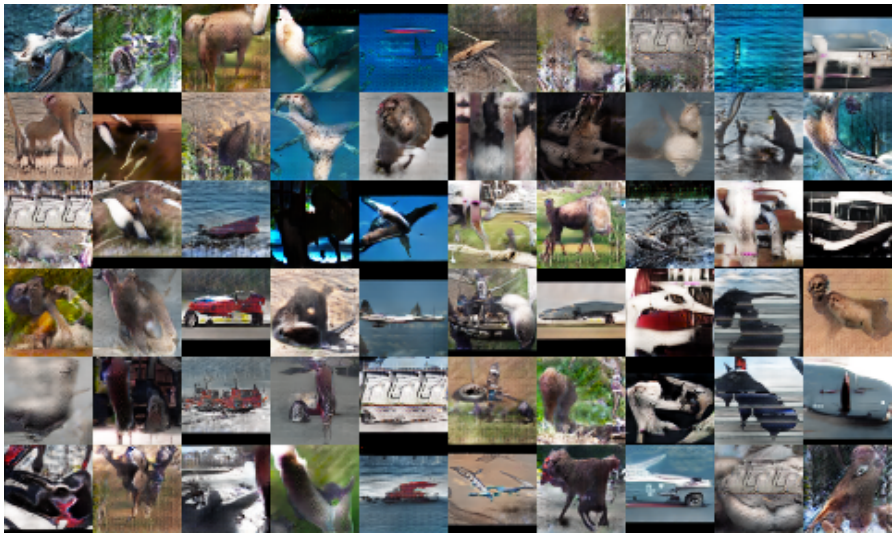

Figure 7. Sample images generated by our best model trained on STL10.

We additionally provide results on the STL-10 dataset, where M-BGAN yields numerical results slightly below Spectral Normalization. Except for the adaptation of the network to $48 \times 48$ images, as done in (Miyato et al., 2018), the experimental setup of the experimental section is left unchanged.

Table 2. Comparison to the state of the art in terms of inception score (IS) and Fréchet inception distance (FID) on the STL-10 dataset.

| Model | IS | FID |
|---|---|---|
| WGP (Miyato et al., 2018) | 8.4 | 55 |
| M-BGAN | 8.7 | 51 |
| SN (Miyato et al., 2018) | 8.7 | 47.5 |
| SN (Hinge loss)(Miyato et al., 2018) | 8.8 | 43.2 |

## 11. M-BGAN as an ensembling method

Intuitively, the M-BGAN loss performs a simple ensembling of many strongly dependant permutation invariant discriminators, at no additional cost.

In the general case, ensembling of N independant discriminators $D_1, \ldots, D_N$ amounts to training each discriminator independently, and using the averaged gradient signal to train the generator. Ensembling is expected to alleviate some of the difficulties of GAN training: as long as one of the discriminators still provides a significant gradient signal, training of the generator is possible. With equation (14), M-BGAN is an ensemble of $B$ permutation invariant discriminators, with respective outputs $1\text{-th}(o_1, \ldots, o_B), \ldots, B\text{-th}(o_1, \ldots, o_B)$, where $i$-th is the function that returns the $i$-th greatest element of a B dimensional vector. Indeed,

$$\frac{1}{N}\sum_{i=1}^{N} l(i\text{-th}(o_1, \ldots, o_B), y) = \frac{1}{N}\sum_{i=1}^{N} l(o_i, y). \tag{54}$$

which is the M-BGAN loss. The ensembled discriminators of the M-BGAN all share the same weights. We believe this ensembling effect at least partially explains the improved performance of M-BGAN.